%% file: iclr2025.tex
\PassOptionsToPackage{table}{xcolor}
\documentclass{article} % For LaTeX2e
\usepackage{iclr2025,times}
\usepackage[utf8]{inputenc}
\usepackage{fancyvrb}

\usepackage{hyperref}
\usepackage{url}
\usepackage{booktabs} % add this in your preamble
\usepackage{float}
\usepackage{xcolor}
\usepackage{booktabs}
\usepackage{caption}
\usepackage{listings}
\usepackage{adjustbox}
\usepackage{microtype}
\usepackage{booktabs}
\definecolor{darkblue}{rgb}{0, 0, 0.5}
\hypersetup{colorlinks=true, citecolor=darkblue, linkcolor=darkblue, urlcolor=darkblue}
\usepackage{wrapfig}
\usepackage{graphicx}
\usepackage{enumitem}
\usepackage{soul}
\usepackage{subcaption}
\usepackage{amsmath}
\usepackage{multicol}   % lets you switch to multiple columns temporarily
\newif\ifcomments

\usepackage[most]{tcolorbox}   % loads xcolor automatically
\definecolor{correctFrame}{HTML}{1B5E20}   % dark green
\definecolor{correctBack}{HTML}{E8F5E9}    % light green
\definecolor{incorrectFrame}{HTML}{B71C1C} % dark red
\definecolor{incorrectBack}{HTML}{FFEBEE}  % light red
\ifcomments
    
    \providecommand{\cmt}[2] % use like \cmt{Alice}{This rocks!}
    {{\protect\color{gray}{[#1: #2]}}}
\else
    \providecommand{\cmt}[2]{}
\fi

\definecolor{queryFrame}{gray}{0.50}
\definecolor{queryBack}{gray}{0.93}

\definecolor{correctFrame}{HTML}{287D3C}
\definecolor{correctBack}{HTML}{EAF5EC}
\definecolor{incorrectFrame}{HTML}{B82020}
\definecolor{incorrectBack}{HTML}{FDEBEC}

\title{FrontierScience: Evaluating AI’s ability to perform expert-level scientific tasks}

\author{%
Miles Wang\thanks{Correspondence to milesw@openai.com} \\
\And Robi Lin \\
\And Kat Hu
\And Joy Jiao \\
\AND Neil Chowdhury \\
\And Ethan Chang \\
\And Tejal Patwardhan \\
\AND \normalfont{OpenAI}
}
\definecolor{A}{HTML}{FFC0C0}      % Pastel Red
\definecolor{Z}{HTML}{FF3131}      % neon Red
\definecolor{B}{HTML}{FFDAC0}        % Pastel Orange
\definecolor{C}{HTML}{FFFDC0}      % Pastel Yellow
\definecolor{D}{HTML}{77dd77}          % Pastel Green
\definecolor{E}{HTML}{C0FFFD}     % Pastel Cyan
\definecolor{F}{HTML}{D0E0FF}            % Pastel Light Blue
\definecolor{G}{HTML}{C0C0FF}             % Pastel Blue
\definecolor{H}{HTML}{DAC0FF}          % Pastel Violet
\definecolor{I}{HTML}{FFC0FF}       % Pastel Magenta
\definecolor{J}{HTML}{FFC0DA}      % Pastel Rose
\definecolor{darkpastelred}{HTML}{C23B22}
\definecolor{darkgreen}{HTML}{1cc650}
\definecolor{lightgreen}{HTML}{caee9c}          % Pastel Green

\definecolor{darkred}{rgb}{0.5,0.0,0.0}
\definecolor{darkyellow}{rgb}{0.7, 0.5, 0.0}

\makeatletter
 \def\SOUL@hlpreamble{%
 \setul{}{2.4ex}%
 \let\SOUL@stcolor\SOUL@hlcolor
 \SOUL@stpreamble
 }
\makeatother

\setlist[itemize]{nosep,leftmargin=*}
\newcolumntype{Y}{>{\raggedright\arraybackslash}X} % table set-up

\definecolor{A}{HTML}{FFC0C0}      % Pastel Red
\definecolor{Z}{HTML}{FF3131}      % neon Red
\definecolor{B}{HTML}{FFDAC0}        % Pastel Orange
\definecolor{C}{HTML}{FFFDC0}      % Pastel Yellow
\definecolor{D}{HTML}{77dd77}          % Pastel Green
\definecolor{E}{HTML}{C0FFFD}     % Pastel Cyan
\definecolor{F}{HTML}{D0E0FF}            % Pastel Light Blue
\definecolor{G}{HTML}{C0C0FF}             % Pastel Blue
\definecolor{H}{HTML}{DAC0FF}          % Pastel Violet
\definecolor{I}{HTML}{FFC0FF}       % Pastel Magenta
\definecolor{J}{HTML}{FFC0DA}      % Pastel Rose
\definecolor{darkpastelred}{HTML}{C23B22}
\definecolor{darkgreen}{HTML}{1cc650}
\definecolor{lightgreen}{HTML}{caee9c}          % Pastel Green

\definecolor{darkred}{rgb}{0.5,0.0,0.0}
\definecolor{darkyellow}{rgb}{0.7, 0.5, 0.0}

\usepackage{pifont}

\everypar{\looseness=-1}

\tcbset{
  plainreddef/.style={
    colback=red!3!white,
    colframe=red!40!black,
    boxrule=0.8pt,
    arc=4pt,
    left=6pt,
    right=6pt,
    top=6pt,
    bottom=6pt
  }
}

\usepackage{tabularx}

\iclrfinalcopy

\begin{document}

\onecolumn
\vspace*{-0.6in} % Moves it lower on the page
\hspace*{1.3in} % Moves it to the right

\maketitle
\normalfont
\vspace*{-0.15in}
\begin{abstract}
% \vspace{-0.5cm}
We introduce FrontierScience, a benchmark evaluating AI capabilities for expert-level scientific reasoning. FrontierScience consists of two tracks: (1) \textbf{Olympiad}, which contains international olympiad problems (at the level of IPhO, IChO, and IBO), and (2) \textbf{Research}, which contains PhD-level, open-ended problems representative of sub-problems in scientific research. In total, FrontierScience is composed of several hundred questions (160 in the open-sourced gold set) covering subfields across physics, chemistry, and biology, from quantum electrodynamics to synthetic organic chemistry. Recent model progress has nearly saturated existing science benchmarks, which often rely on multiple-choice knowledge questions or already published information. In contrast, all Olympiad problems are originally produced by international olympiad medalists and national team coaches to ensure standards of difficulty, originality, and factuality. All Research problems are research sub-tasks written and verified by PhD scientists (doctoral candidates, post-doctoral researchers, or professors). For Research, we also introduce a granular rubric-based architecture to evaluate model capabilities throughout the process of solving a research task, as opposed to judging a standalone answer. In initial evaluations of several frontier models, GPT-5.2 is the top performing model on FrontierScience, scoring 77\% on the Olympiad set and 25\% on the Research set.
\end{abstract}

\vspace{-0.2cm}
\section{Introduction}\label{sec:intro}\input{sections/01-intro}
\section{Benchmark Construction}\label{sec:benchmark}\input{sections/02-method}
\section{Experiments}\label{sec:experiments}\input{sections/03-datasets}
\section{Discussion}\label{sec:discussion}\input{sections/06-conclusion}
\section{Related Work}\label{sec:related}\input{sections/07-related}

\bibliography{conference}
\bibliographystyle{iclr2025}
% APPENDIX
\clearpage
\appendix
\section*{Appendix}\label{sec:appendix}\input{sections/99-appendix}
% \addcontentsline{toc}{section}{Appendix} % Optional: add to TOC

\end{document}

%% file: sections/01-intro.tex
Language models’ reasoning capabilities have significantly advanced in scientific domains. When GPQA, a “Google-Proof” multiple-choice science benchmark written by PhD experts, was released in November 2023, GPT-4 scored 39\%, below the expert baseline of 70\% \citep{rein2023gpqa}. Two years later, GPT-5.2 scored 92\% \citep{openai2025gpt5p2}.

As models’ reasoning and knowledge capabilities continue to scale, unsaturated benchmarks will be important to measure and forecast models’ ability to accelerate scientific research. Prior benchmarks have tracked useful scientific capabilities relative to model improvements \citep{rein2023gpqa, he2024olympiadbench, lu2022scienceqa, hendrycks2021mmlu}. However, as models have rapidly improved at reasoning, a new generation of science benchmarks is required to keep apace with progress.

% they focus on multiple-choice questions rather than open-ended reasoning tasks \citep{rein2023gpqa}, collect online datasets that may be contaminated during model training \citep{he2024olympiadbench}, prioritize high-school level questions \citep{lu2022scienceqa,hendrycks2021mmlu}, or are not primarily focused on science \citep{phan2025hle,glazer2024frontiermath}.

To assess real-world scientific capabilities, we introduce FrontierScience, composed of hundreds of questions that are difficult, verifiable, and original. FrontierScience questions are written and verified by subject matter experts across physics, chemistry, and biology, and are composed of two levels:
\begin{enumerate}
    \item \textbf{FrontierScience-Olympiad}: Science Olympiad-style questions designed by international olympiad medalists to assess scientific reasoning in a short answer format.
    \item \textbf{FrontierScience-Research}: Research subproblems designed by PhD scientists (doctoral candidates, professors, or postdoctoral researchers) that one might encounter while doing original research.
\end{enumerate}

\begin{figure}[t!]
    \centering
    \includegraphics[width=\linewidth]{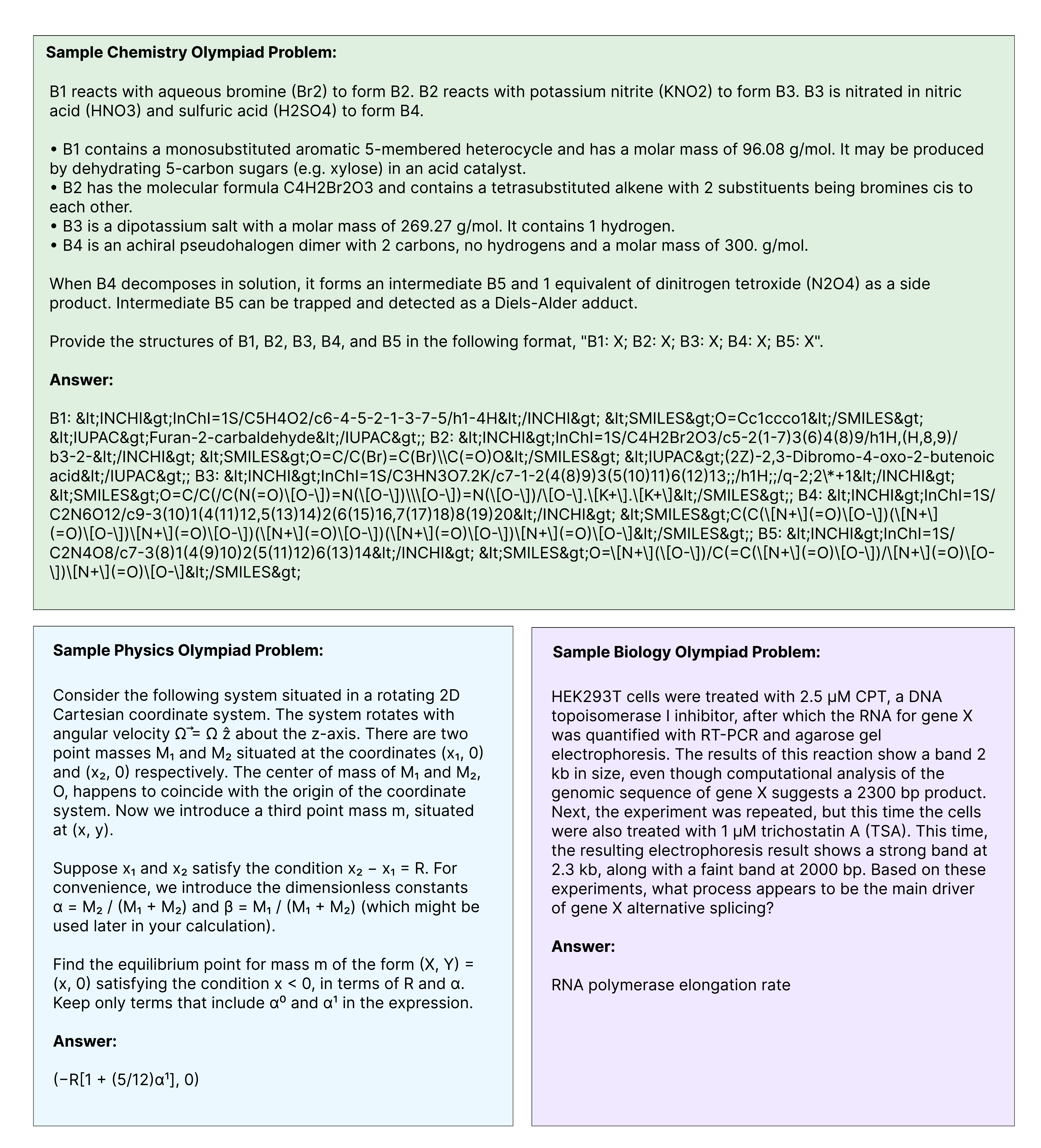}
    \caption{\textbf{Sample FrontierScience-Olympiad problems.} Each task in FrontierScience is written and verified by a domain expert in physics, chemistry, or biology. For the Olympiad set, all experts achieved a medal in an international olympiad competition.}
    \label{fig:example_problem_figure1}
\end{figure}

%  For the Research set, all experts hold a relevant PhD degree

We constructed this dual evaluation set\footnote{Dataset: \url{https://huggingface.co/datasets/openai/frontierscience/tree/main}} to measure two sets of capabilities. The Olympiad set is designed to evaluate precise problem solving in a constrained setting. The problems are designed such that solutions can be evaluated with a single numeric or algebraic expression (physics and chemistry) or a fuzzy string-matchable answer (biology). The Research set evaluates more open-ended reasoning, judgment, and the ability to support real world research objectives. Each Research problem is accompanied by an expert-designed, 10-point rubric. Together, they provide a wider diagnostic of model strengths and weaknesses for expert-level scientific reasoning than previous benchmarks.

In initial evaluations of several frontier models, GPT-5.2 is the overall top performing model on FrontierScience, scoring 77\% on the Olympiad set and 25\% on the Research set. Gemini 3 Pro is comparable to GPT-5.2 on Olympiad, scoring 76\%, and GPT-5 ties GPT-5.2 on Research at 25\%. Overall, we find that frontier AI systems have rapidly progressed in solving expert-level reasoning questions, particularly at the level of self-contained olympiad problems, but are still far from saturation on research-style work.

\begin{figure}[t!]
    \centering
    \includegraphics[width=\linewidth]{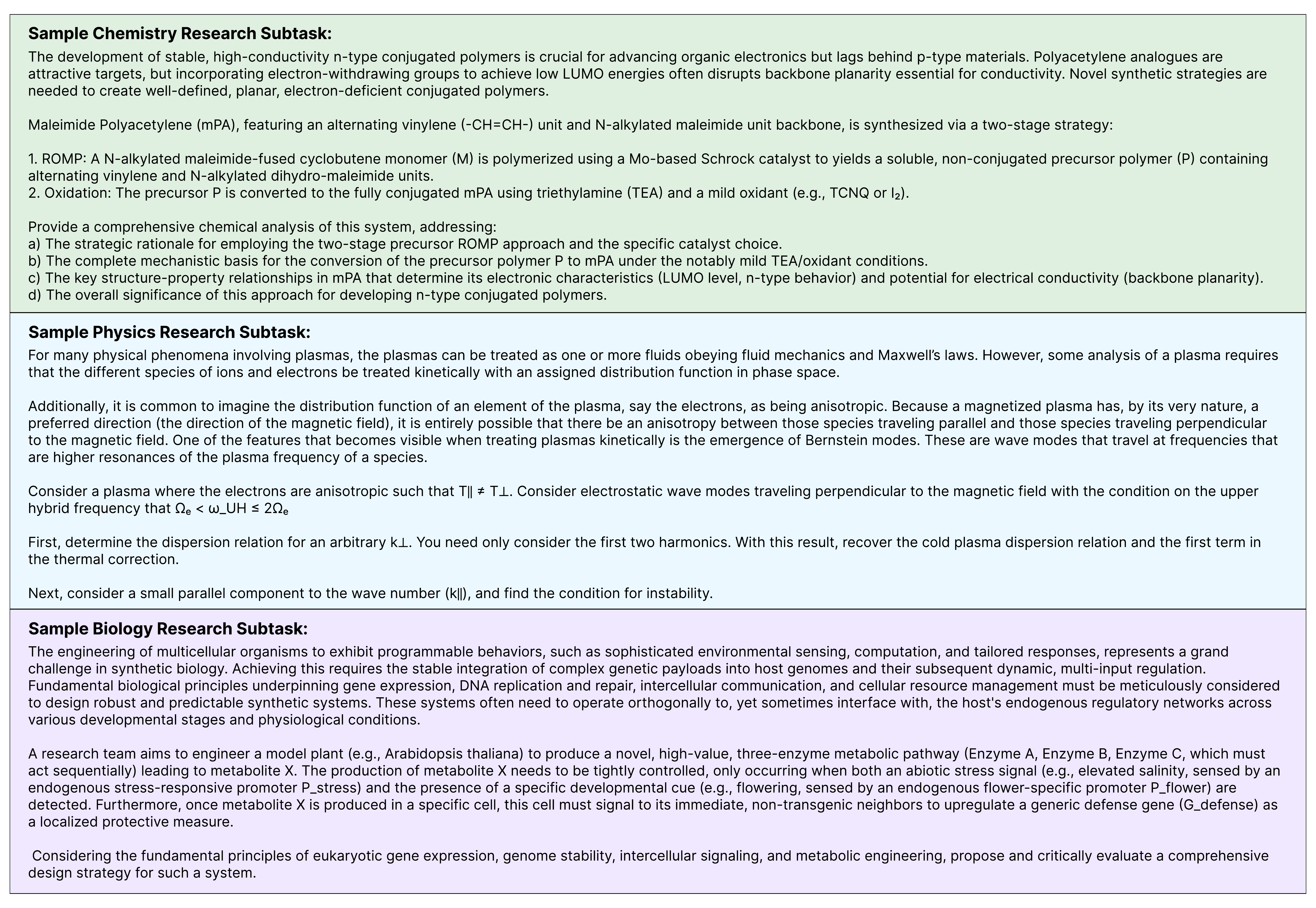}
    \caption{\textbf{Sample FrontierScience-Research problems.} For the Research set, all experts hold a relevant PhD degree. The corresponding rubrics to these sample tasks can be found in Appendix \ref{appendix:samples}.}
    \label{fig:example_problem_figure1}
\end{figure}

%% file: sections/02-method.tex
\subsection{Data collection pipeline}
\label{sec:pipeline}

FrontierScience-Olympiad questions were created in collaboration with 42 former international medalists or national team coaches in physics, chemistry, or biology, who have achieved 108 olympiad medals in total (45 gold, 37 silver, 26 bronze). All medalists were awarded from at least one (and often multiple) of the following olympiads: the International Physics Olympiad, International Chemistry Olympiad, International Olympiad on Astronomy and Astrophysics, European Physics Olympiad, and International Mendeleev Chemistry Olympiad.

FrontierScience-Research questions were created in collaboration with 45 qualified scientists and domain experts. The scientists were either post-doctoral researchers, professors, or doctoral candidates, often from globally recognized institutions. Qualitatively, each task was designed to represent a subproblem a PhD researcher might need to solve during the course of their research, and take at least three to five hours to successfully complete.

The scientists' areas of expertise spanned an array of scientific disciplines, including but not limited to: quantum mechanics, astrophysics, theoretical and experimental physics, biophysics, nanotechnology; molecular, evolutionary, and developmental biology, pharmacology, genomics, immunology, and neuroscience in biology; and biochemistry, physical and organic chemistry, materials and computational chemistry, catalysis, and photochemistry in chemistry. Experts are actively engaged in research for their domain with deep familiarity of research methodologies. 

Each scientist wrote original problems for their track adhering to the following guidelines:

\begin{tabularx}{\linewidth}{@{} >{\bfseries\raggedright\arraybackslash}l *{2}{Y} @{}}
\toprule
 & \textbf{Olympiad} & \textbf{Research} \\
\midrule
Originality &
\begin{itemize}[leftmargin=*,nosep]
  \item Problems are designed to mimic olympiad style challenges of complex, closed-form reasoning tasks.
  \item To minimize contamination risks, all problems are novel. While problems could draw initial inspiration from existing scientific ideas or questions, problems need to be re-contextualized for measuring reasoning capabilities through creative and non-obvious combinations or modifications.
\end{itemize}
&
\begin{itemize}[leftmargin=*,nosep]
  \item Problems are designed to represent authentic scientific research tasks grounded in active areas of inquiry by our contributors. The specific problems will not be published outside of this dataset.
  \item To minimize contamination risks, all problems are expected to be novel. For problems that draw on the same core phenomena as published areas of study, or inspired by other problems, originality is enforced through the review process to ensure no direct overlap.
\end{itemize}
\\
\midrule
Difficulty &
\begin{itemize}[leftmargin=*,nosep]
  \item Problems at least the level of difficulty of international olympiad questions.
  \item Preliminary questions were evaluated against various internal models, where if the model answered correctly, the question was considered invalid and required an update.
\end{itemize}
&
\begin{itemize}[leftmargin=*,nosep]
  \item Preliminary questions were evaluated against various internal OpenAI models. If the model scored highly on the rubric, the question was either discarded or significantly modified.
  \item Problems were calibrated such that 7-8 points out of 10 on the rubric was considered a successful solution.
\end{itemize}
\\
\midrule
Verifiability &
\begin{itemize}[leftmargin=*,nosep]
  \item The question provides all necessary variables, units, and information used in the final answer.
  \item The answer should be a single numeric or algebraic expression (physics and chemistry) or a fuzzy string-matchable answer (biology).
\end{itemize}
&
\begin{itemize}[leftmargin=*,nosep]
  \item Each question includes a scoring rubric with multiple independent and objectively assessable items, totaling 10 points.
  \item Model judgments did not meaningfully differ from human judgments when using the rubric.
\end{itemize}
\\
\bottomrule
\end{tabularx}

For each research and olympiad problem, scientists provided a detailed solution that would earn full credit, as well as associated metadata (subdomains, difficulty levels, and sources of inspiration). Each contributed problem then went under review by at least one peer domain expert (for Research, each problem went under at least two reviews), who evaluated all components of the question against the guidelines. Questions could be inspired by known problems, or reference past work, but the guidelines were to make the task still new.

% We publish XX\% of the dataset, holding out the other YY\% to track potential contamination or overfitting risks.

\subsection{Verification pipeline}
\label{sec:verification}

\begin{figure}[t]
    \centering
    \includegraphics[width=\linewidth]{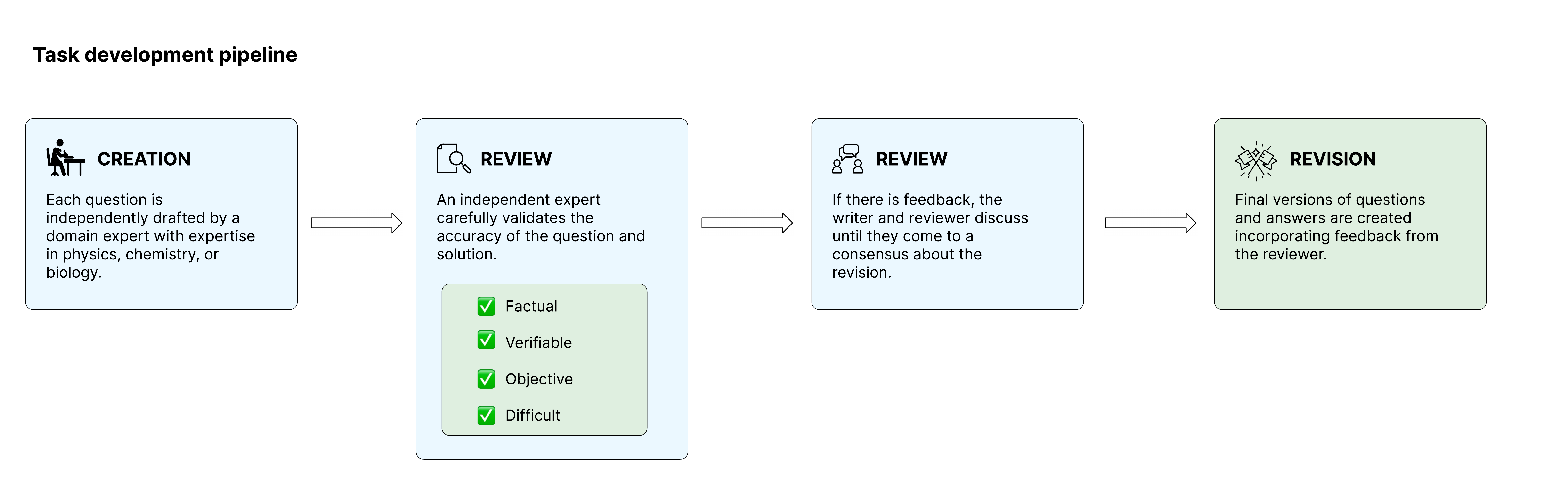}
    \caption{Tasks go through four stages: Creation, Review, Resolution, Revision. Independent experts review each other tasks to verify it aligns with the criteria.}
    \label{fig:verification}
\end{figure}

All submitted questions underwent an iterative review process. Independent domain expert reviewers read through each question, answer (either short answer for Olympiad, or rubric for Research), and solution explanation. The reviewers verified that each question was correct and followed all of the guidelines. The task creation process included some selection against OpenAI internal models (e.g., discarding tasks that models successfully got right, so we expect the evaluation to be biased against these models relative to others.

If any disagreements arose between the question writer and reviewer, they either came to a consensus or the question was discarded. Only after both experts agreed was the question submitted and added to the dataset. Experts for each domain in set then did a final review over each question in the submitted dataset, ensuring that all questions aligned with the guidelines.

For the Olympiad set, all problems went through at least one independent review, and then a holistic review by experts. For the Research set, all problems went through at least two independent reviews, and then a meta review by the experts. We increased review coverage for Research due to the questions being open-ended and rubrics being a newer and more imprecise grading architecture.

From over 500 Olympiad questions and over 200 Research questions, we did a meta-review with experts to filter down to an open-sourced gold set of 100 Olympiad questions and 60 Research questions. We keep the rest of the questions held-out to track potential contamination of the open-sourced set.

\subsection{Rubric-based grading}
\label{sec:rubric}

\begin{figure}[t]
    \centering
    \includegraphics[width=\linewidth]{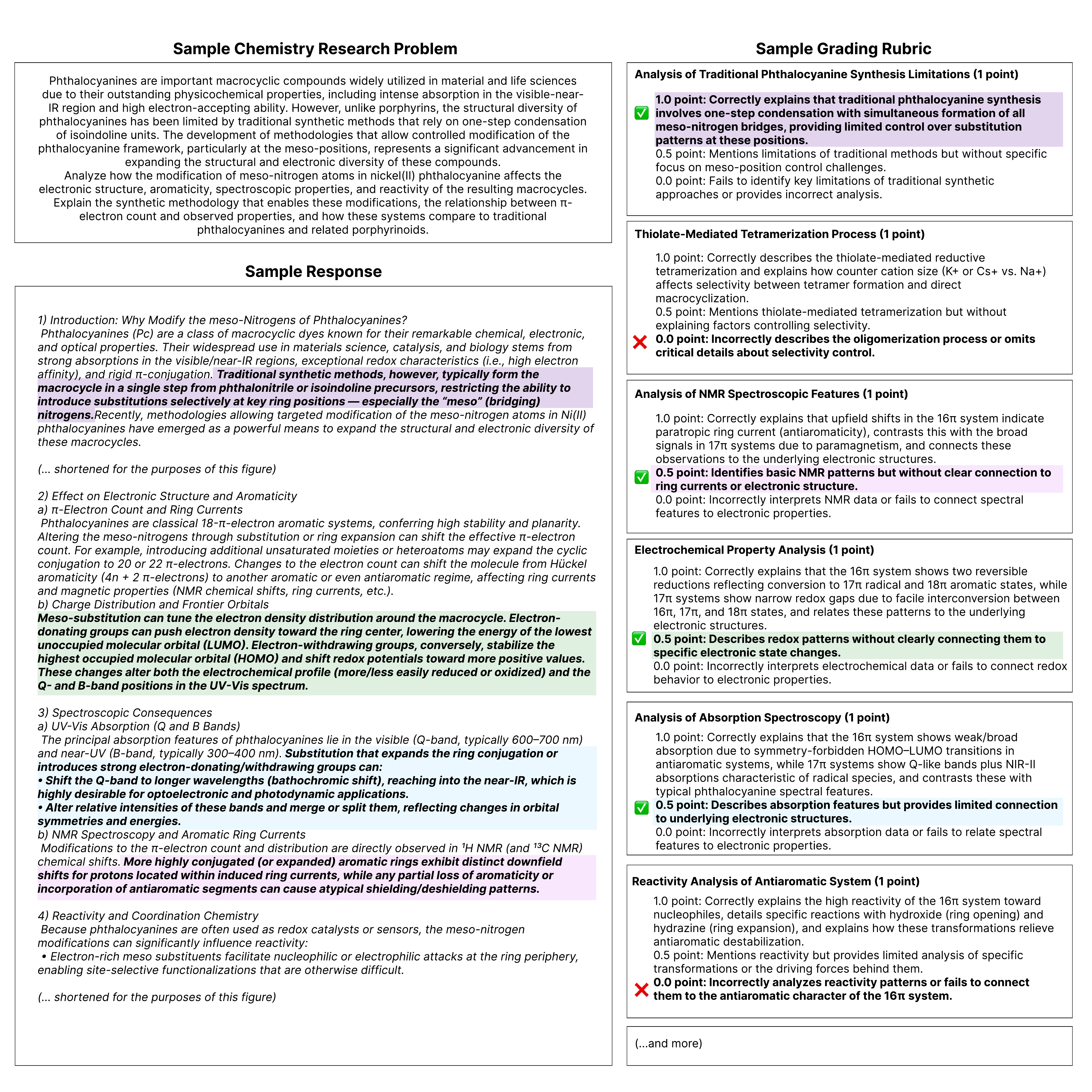}
    \caption{Each task in the research set is graded using a rubric totaling 10 points that can be used by an expert or a judge model. To scale our ability to evaluate models, we use another model to judge responses.}
    \label{fig:rubric}
\end{figure}

The Olympiad set is gradable with a number, expression, or fuzzy string match, which improves verification. However, this verification often trades off with the expressivity and open-endedness of the problem. For the Research set, we introduce an experimental rubric-based architecture for grading more open-ended tasks.

Each question includes a scoring rubric with multiple independent and objectively assessable items, totaling 10 points. Each rubric item contains a description for a specific pass/fail condition (e.g., “Writes the following equation X”) and points. The grading rubric assesses not only the accuracy of the final answer, but also the correctness of intermediate reasoning steps, allowing for nuanced model performance and failure analysis. Scoring seven out of 10 points is considered a suitable solution and marked as a success. Due to the experimental design, we expect the Research set to have a lower noise ceiling than the Olympiad set. The flexibility of rubric points also enables other future grading procedures, such as average rubric points or different thresholds for what is considered a "success".

Each question is accompanied by an explanatory solution path crafted by subject-matter experts. To run these evaluations without requiring human expert graders, we rely on judge models that assign a score given an attempted answer and a rubric. We provide model judge prompts in Appendix \ref{appendix:eval_prompts} that we use for all evaluations in this paper. We use GPT-5 at high reasoning effort for the model judge.

\subsection{Benchmark composition}
\label{sec:composition}
\begin{figure}[t]
    \centering
    \includegraphics[width=0.6\linewidth]{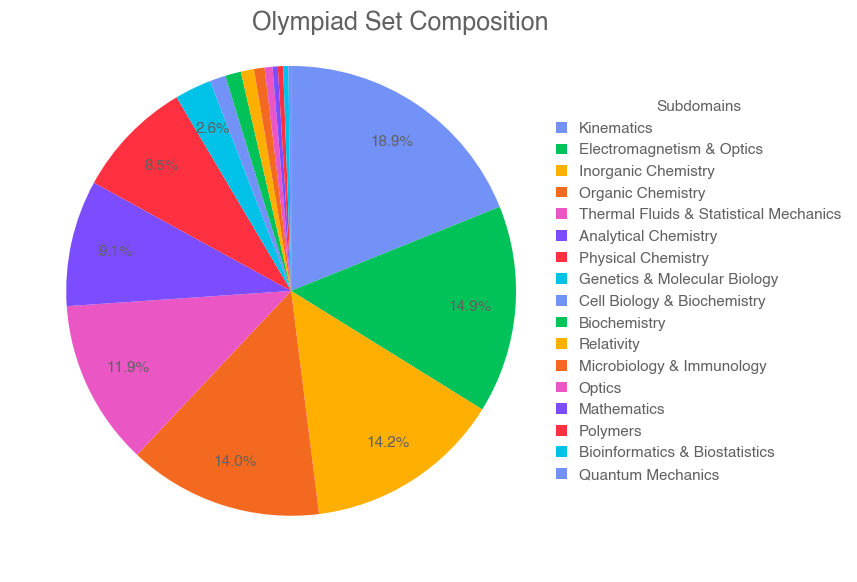}
    \caption{The Olympiad split is composed of a diverse set of topics, from biochemistry to quantum mechanics.}
    \label{fig:composition}
\end{figure}

FrontierScience contains a diverse range of scientific questions (Fig.~\ref{fig:composition}). The Olympiad set is grounded in topics common on international science olympiad exams and is more weighted toward physics and chemistry over biology because it’s more feasible to develop questions that resolve to verifiable expressions and numbers. The Research set is grounded in contributors’ research specialties, with the gold set of 60 questions equally split between physics, chemistry, and biology.

%% file: sections/03-datasets.tex
\subsection{Main results}
\label{sec:main_results}

% \begin{figure}[h]
%     \centering
%     \includegraphics[width=0.8\linewidth]{figures/headline_plot.png}
%     \caption{We compare accuracies across several frontier models. GPT-5 is our highest performing model on the FrontierScience-Research set and is tied with Grok 4 and Gemini 2.5 Pro on the Olympiad set. TODO: Update this plot to the one in the blog post.}
%     \label{fig:headline}
% \end{figure}

\begin{figure}[h]
    \centering
    \begin{subfigure}[t]{0.48\linewidth}
        \centering
        \includegraphics[width=\linewidth]{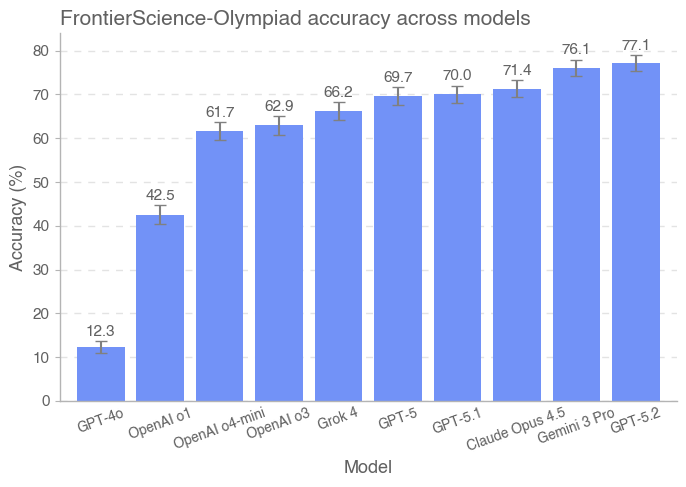}
    \end{subfigure}\hfill
    \begin{subfigure}[t]{0.48\linewidth}
        \centering
        \includegraphics[width=\linewidth]{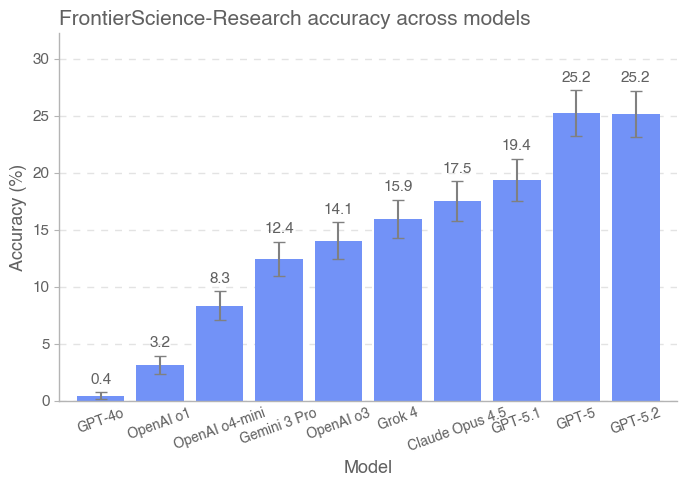}
    \end{subfigure}
    \caption{We compare accuracies across several frontier models. GPT-5.2 is our highest performing model across the Olympiad and the Research set. Gemini 3 Pro is comparable to GPT-5.2 on Olympiad, and GPT-5 is tied with GPT-5.2 on Research. For all Olympiad evaluations, scores were averaged across 20 independent trials. For all Research evaluations, scores were averaged across 30 independent trials, using a threshold of a response earning at least seven rubric points as correct.}
    \label{fig:headline}
\end{figure}

We evaluated several frontier models: GPT-4o, OpenAI o4-mini, OpenAI o3, GPT-5.2, Claude Opus 4.5, Gemini 3 Pro, Grok 4, GPT-5.1, and GPT-5 on FrontierScience-Olympiad and FrontierScience-Research. All reasoning models were evaluated at “high” reasoning effort with the exception of GPT-5.2 at “xhigh”, and without browsing. In our initial evaluations, GPT-5.2 is the top performing model on FrontierScience, scoring 77\% on the Olympiad Set and 25\% on the Research set. Surprisingly, GPT-5 outperforms GPT-5.1 on the Research set and ties GPT-5.2. Overall, we’ve seen substantial progress on solving expert-level questions while leaving headroom for more progress, especially on open-ended research-style tasks. 

For both sets, we use a model-based judge of GPT-5 (with a reasoning effort of "high") to evaluate the models. For Olympiad, we give the judge the attempted answer and the actual answer and ask it to compare equivalency of the expression, number, or phrase. For Research, we give the judge the attempted answer and the rubric and ask it to return a single number reflecting the number of rubric points the answer earned (Appendix \ref{appendix:eval_prompts}).

\begin{figure}[h]
    \centering
    \begin{subfigure}[t]{0.48\linewidth}
        \centering
        \includegraphics[width=\linewidth]{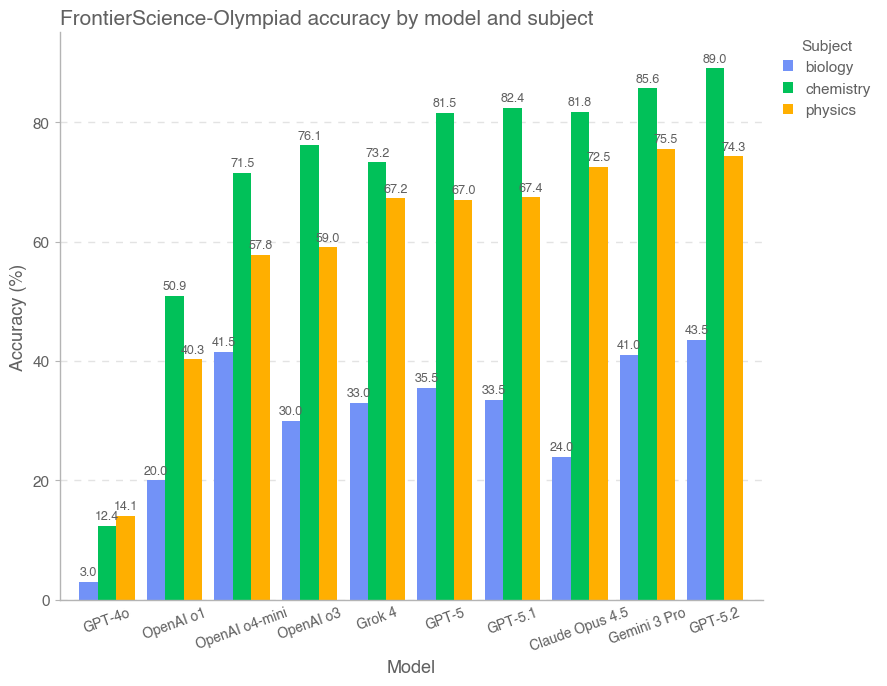}
    \end{subfigure}\hfill
    \begin{subfigure}[t]{0.48\linewidth}
        \centering
        \includegraphics[width=\linewidth]{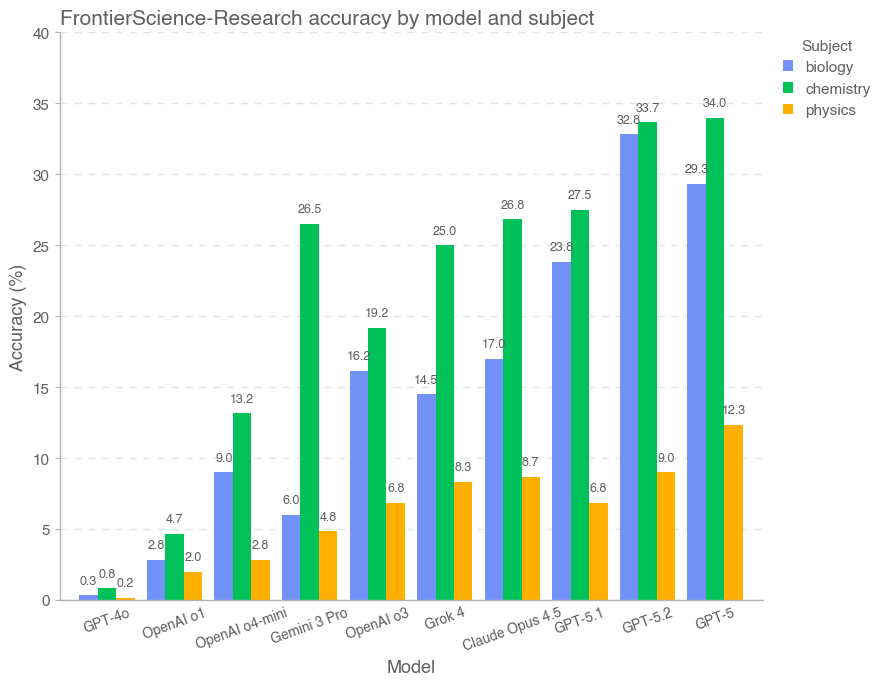}
    \end{subfigure}
    \caption{We compare accuracies across several frontier models on FrontierScience-Olympiad (Left) and FrontierScience-Research (Right) with accuracies split out by subject.}
    \label{fig:headline}
\end{figure}

When separating by subject, models perform comparably across distributions. For the Olympiad set, models perform better on chemistry, followed by physics and biology. For the Research set, models perform better on chemistry, followed by biology and then physics. Analyzing the transcripts, models typically struggled with reasoning or logic error, failures in understanding niche concepts, calculation errors, and factual inaccuracy.

% \subsection{Increasing test-time compute}
% \label{sec:pass_rates}

\begin{figure}[h]
    \centering
    \begin{subfigure}[t]{0.48\linewidth}
        \centering
        \includegraphics[width=\linewidth]{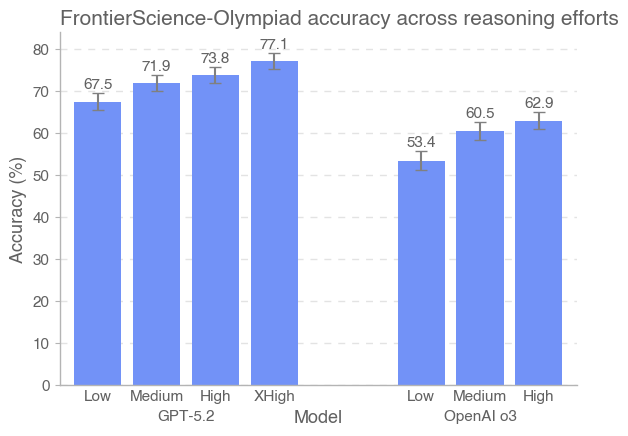}
    \end{subfigure}\hfill
    \begin{subfigure}[t]{0.48\linewidth}
        \centering
        \includegraphics[width=\linewidth]{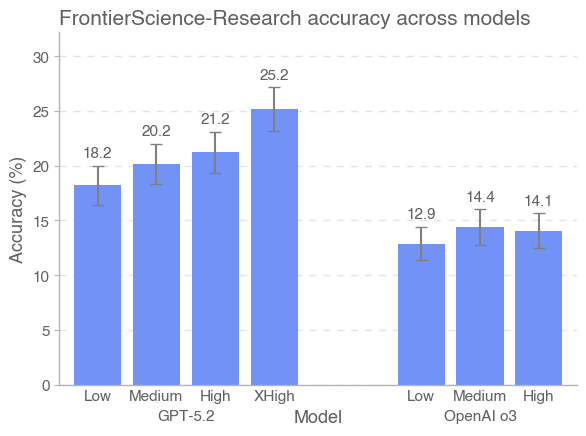}
    \end{subfigure}
    \caption{We compare accuracies for GPT-5.2 and OpenAI o3 on FrontierScience-Olympiad and FrontierScience-Research across different reasoning efforts. Using more test-time tokens enables GPT-5.2 to go from 67.5\% to 77.1\% on the Olympiad set, as well as from 18\% to 25\% on the Research set. Surprisingly, o3 performs marginally worse at high reasoning effort compared to medium reasoning effort on the Research set.}
    \label{fig:headline}
\end{figure}

% We evaluate the models with a pass@k metric (Chen et al., 2021), where we give the model k tries and estimate if it would have succeeded on any of the k tries. We find a steady increase when evaluating up to pass@10 on the Olympiad and Research sets, with o1 going from 56\% to ~80\% on Olympiad and from 6\% to 22\% on Research.

% \subsection{Pass rates analysis}
% \label{sec:pass_rates}

% \begin{figure}[h]
%     \centering
%     \includegraphics[width=0.8\linewidth]{figures/pass_rates.png}
%     \caption{TODO: caption, ALSO DO THIS FOR RESEARCH"}
%     \label{fig:pass_rates}
% \end{figure}

% We also inspect the pass rates distribution. Most Olympiad problems have been solved at least one time out of 10 tries by o3, although a significant chunk have never been solved. The vast majority of Research problems have never been solved by o3.

% \subsection{Rubric scores}
% \label{sec:rubric_scores}
% While we treat a Research solution as solved if it scores at least 8/10 rubric points, we can use the rubric scores as a more continuous metric of progress. When plotting the average rubric points earned, we see a larger gap between models: o3 scores an average of ~4 points per question, while GPT-4o scores ~1.6 points per question. The rubric score distribution for o3 is centered around 3-4 rubric points.

% \begin{figure}[h]
%     \centering
%     \includegraphics[width=0.8\linewidth]{figures/rubric_score.png}
%     \caption{TODO: caption, ALSO DO THIS FOR RESEARCH}
%     \label{fig:rubric_score}
% \end{figure}

%% file: sections/06-conclusion.tex
While FrontierScience represents an advance in understanding scientific research capabilities, there are multiple limitations:

\begin{enumerate}
    \item \textbf{Constrained problem-solving}: A significant part of scientific research is proposing novel research directions, hypotheses, and ideas. FrontierScience is composed of questions with a constrained problem statement, which focuses on evaluating the reasoning to complete a research task and less on ideation. While the Research set aims to measure more open-ended reasoning, this is an inherent limitation of an autogradable Q\&A style evaluation.
    \item \textbf{Rubric reliability}: We sought to improve rubric reliability of the Research set through strict guidelines, verification, and consistency with human grading. However, the rubric is less objective than the equivalency checker of a single expression or number, and relies on the model judge’s capabilities.
    \item \textbf{Modalities}: Problems are designed to be text-only without image or video outputs. Modalities beyond text are more representative of scientific research. In particular, real-world scientific research often involves interaction with reality (e.g., wet labs), which this evaluation does not cover.
    \item \textbf{Human baselining}: We did not perform human baselining of this dataset and leave that to future work. Since the questions are grounded in experts’ authentic research, an interesting question is how to conduct a human baseline. Since the questions are so specialized, it may be important to find experts in that speciality to solve them and derive a consensus baseline.
\end{enumerate}

Research and practical evaluations will be important to continue building long standing and directly relevant evaluations. Scientific reasoning is important for the beneficial impacts of AI and we hope for continued development of robust and relevant benchmarks for accelerating scientific progress.

%% file: sections/07-related.tex
Previous science-oriented and knowledge benchmarks primarily measure models' capabilities through multiple-choice or single-answer formats. Benchmarks such as MMLU \citep{hendrycks2021mmlu, wang2024mmluprorobustchallengingmultitask}, GPQA \citep{rein2023gpqa}, and ScienceQA \citep{lu2022scienceqa} have significantly contributed to understanding model performance across scientific knowledge and basic reasoning tasks. However, these benchmarks largely target knowledge retrieval or recognition of well-known scientific concepts rather than research-level scientific reasoning. GPQA, for instance, measures general-purpose science reasoning but is limited to structured multiple-choice settings, reducing diagnostic power of more complex, open-ended tasks.

To evaluate more advanced reasoning skills, recent benchmarks have introduced open-ended questions. OlympiadBench \citep{he2024olympiadbench} introduced high-school-level Science Olympiad-style questions, demonstrating the value of open-ended and auto-gradable formats. However, it is focused on collecting pre-existing math and physics questions, raising contamination concerns. The FrontierScience Olympiad track extends this format by employing international Olympiad medalists across a range of scientific subjects to craft problems specifically adversarial against state-of-the-art models. Complementary work such as PHYBench \citep{qiu2025phybenchholisticevaluationphysical}, ChemBench \cite{mirza2024largelanguagemodelssuperhuman}, and SciBench \citep{wang2024scibenchevaluatingcollegelevelscientific} also extend constrained reasoning tasks to certain domains. Other benchmarks, such as PaperBench \citep{starace2025paperbenchevaluatingaisability} investigate capabilities on AI research tasks for replicating papers, while being less focused on scientific capabilities.

CritPt \citep{zhu2025probingcriticalpointcritpt} introduces a PhD-level physics benchmarks that focuses on difficult, unpublished research questions, employing a methodology of verifiable checkpoints. FrontierScience trades off on the benefits of fully verifiable checkpoints to evaluate more open-ended research subtasks, which is also shown by its extension to chemistry and biology questions. LAB-Bench \citep{laurent2024labbenchmeasuringcapabilitieslanguage} is a broad and diverse benchmark of biology questions that are relevant to practical workflows. It focuses on multiple choice questions across skills such as recalling literature and manipulating DNA and protein sequences. FrontierScience is complementary to this work by focusing on difficult reasoning questions rather than day-to-day scientific workflows.

Previous approaches to open-ended problem evaluation typically rely on final-answer correctness as a primary assessment metric, limiting the insight into intermediate reasoning steps. Prior work such as the LLM-Rubric \citep{Hashemi_2024} have incorporated rubric-based evaluations for evaluating LLM responses in dimensions of naturalness, conciseness, etc, and recent work such as HealthBench \citep{arora2025healthbenchevaluatinglargelanguage} has used this format in real-world relevant domains. The FrontierScience Research benchmark builds upon the structured rubric-based evaluation approach to test for reasoning with custom rubric items per evaluation. Each rubric item in FrontierScience Research problems explicitly decomposes answers into granular components, enabling more nuanced analyses of where and why models succeed or fail, particularly valuable given the complexity of PhD-level scientific research tasks.

\section{Acknowledgments}

We sincerely thank our external partners and expert evaluators for their valuable contributions, including their time, domain expertise, and thoughtful feedback.

We thank Addea Gupta, Alex Karpenko, Andy Applebaum, Bowen Jiang, David Robinson, Elizabeth Proehl, Evan Mays, Grace Kim, Ilge Akkaya, Jerry Tworek, Joy Jiao, Kevin Liu, Leon Maksin, Leyton Ho, Michele Wang, Nat McAleese, Nikolai Eroshenko, Olivia Watkins, Patrick Chao, Phillip Guo, Phoebe Thacker, Rahul Arora, Ryan Kaufman, Samuel Miserendino, Sebastian Bubeck, Simón Fishman, Stephen McAleer, and Ven Chandrasekaran for helpful discussions, feedback, and support.

%% file: sections/99-appendix.tex
\section{Full sample research problems}
\label{appendix:samples}

\begin{figure}[h!]
    \centering
    \includegraphics[width=\linewidth]{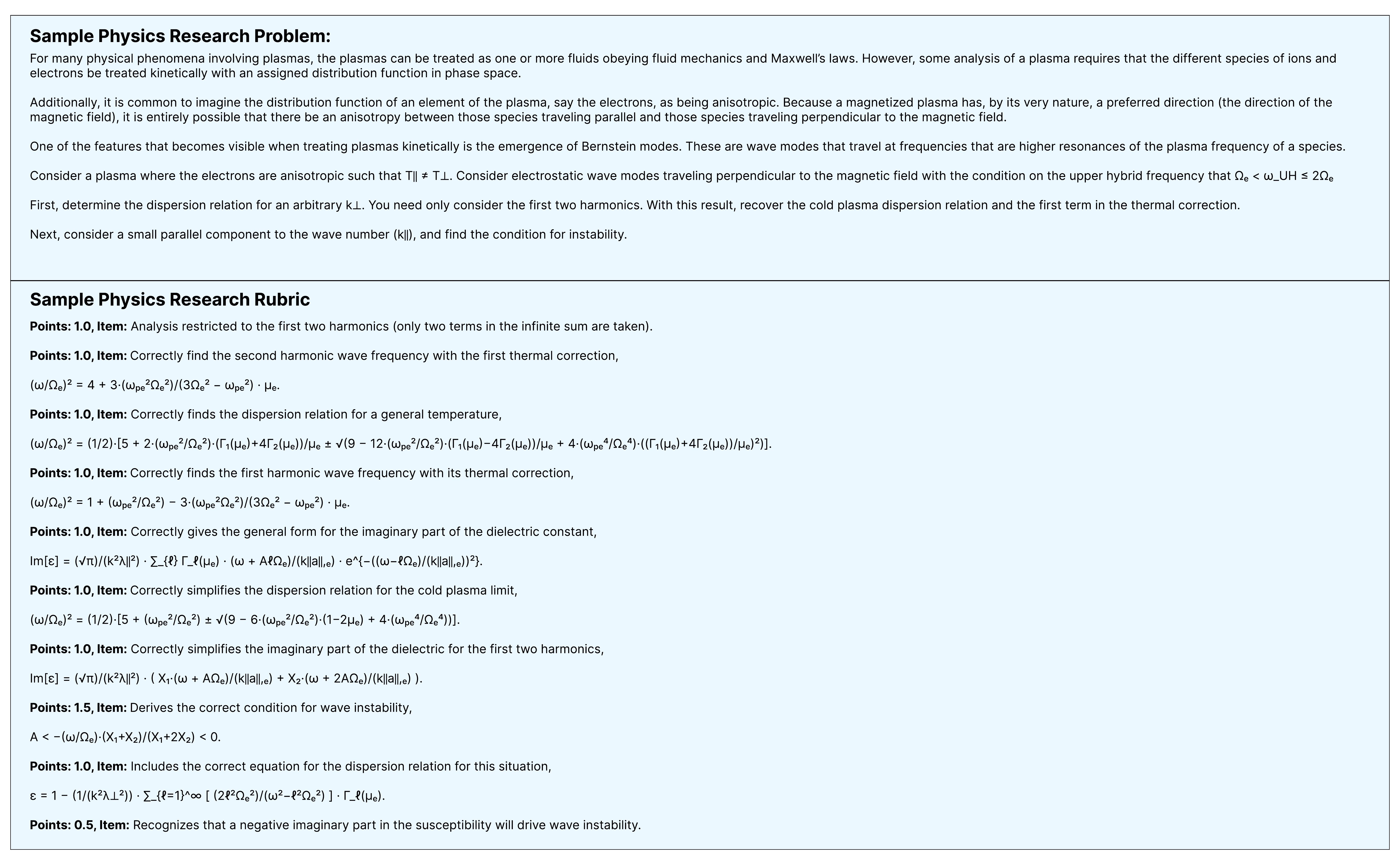}
    \caption{\textbf{Sample FrontierScience-Research Physics Problem.}}
    \label{fig:example_physics_research_problem}
\end{figure}

\begin{figure}[h!]
    \centering
    \includegraphics[width=\linewidth]{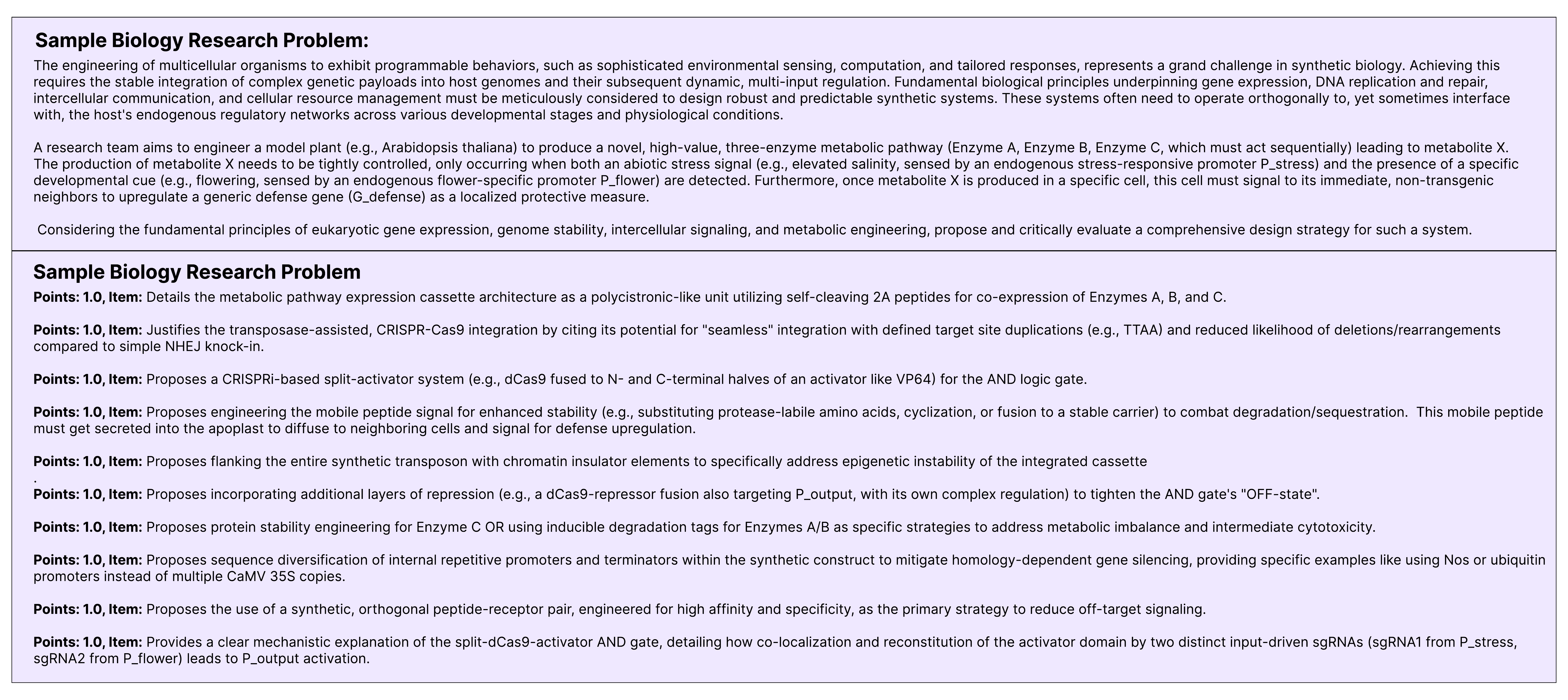}
    \caption{\textbf{Sample FrontierScience-Research Biology Problem.}}
    \label{fig:example_biology_research_problem}
\end{figure}

\begin{figure}[h!]
    \centering
    \includegraphics[width=\linewidth]{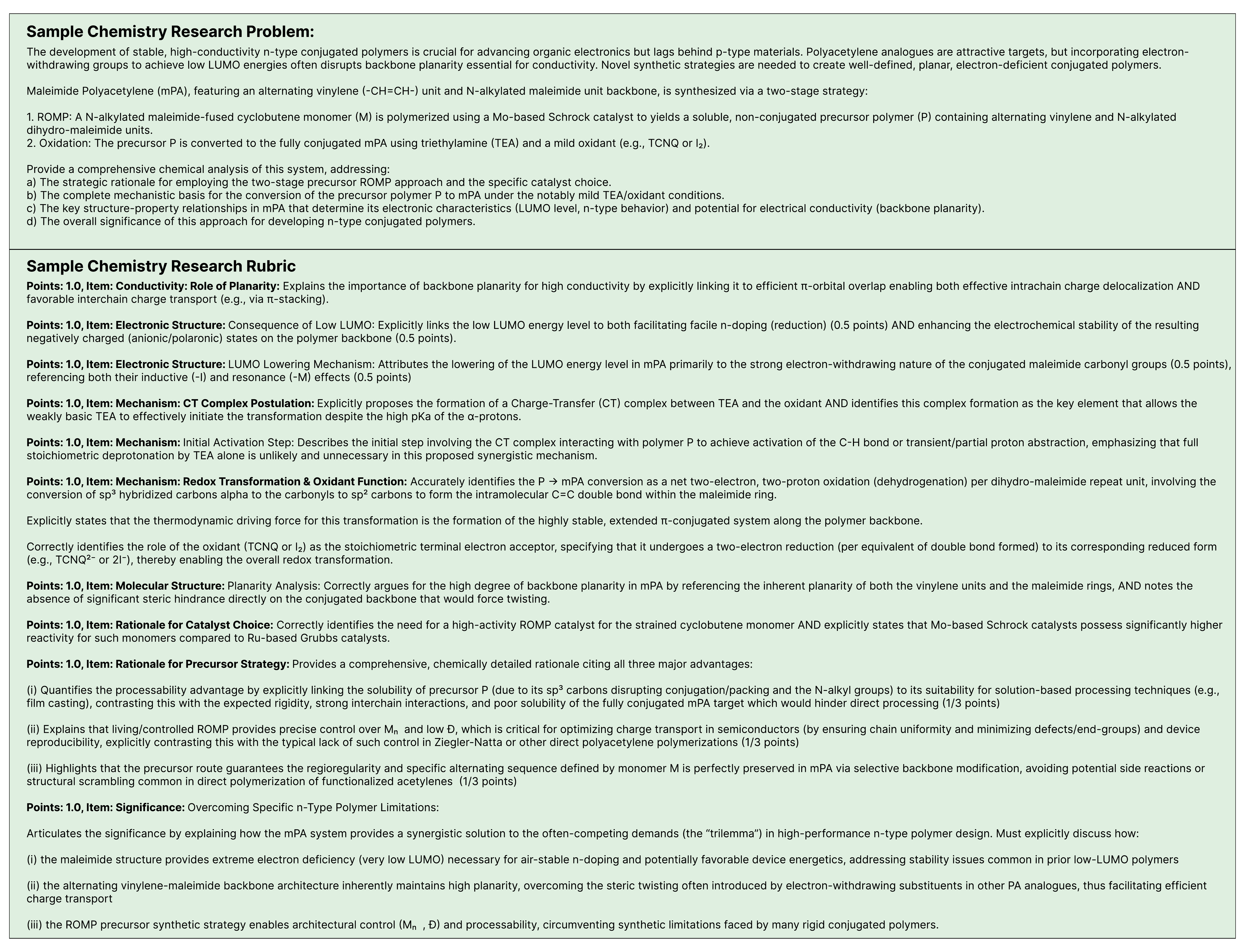}
    \caption{\textbf{Sample FrontierScience-Research Chemistry Problem.}}
    \label{fig:example_chemistry_research_problem}
\end{figure}

\section{Evaluation prompts}
\label{appendix:eval_prompts}
For our evaluations, we use a judge model based on GPT-5 thinking at high reasoning effort to judge responses. Here, we provide the exact prompts we give the judge model for the evaluations in this paper.

\newtcolorbox{shadedbox}[1]{%
  colback=gray!10,
  colframe=gray!60,
  boxrule=0.5pt,
  arc=2mm,
  title=#1,
  fonttitle=\bfseries,
  coltitle=black,
  left=1em, right=1em,
  top=0.8em, bottom=0.8em,
  % �� add this line:
  before upper={\setlength{\parskip}{0.6em}},
}

\begin{shadedbox}{FrontierScience-Olympiad Judge Model Prompt}
You are grading an attempted answer to a science olympiad problem. You will be given the problem, attempted answer, and reference answer. Evaluate the solution against the provided reference solution, ensuring it is complete and matches the reference solution. Pay close attention to detail and grade it strictly, but fairly.

The reference answer is either a single number or expression in latex formatting, a chemical formula, a compound name, or a phrase referring to a specific name, entity, or method.

Mark the attempted answer as correct if it fully matches the reference answer or is otherwise equivalent (e.g., an equivalent algebraic expression, a numerical number within 1 decimal place rounding of the reference answer (e.g., $6.69 \approx 6.7$), an equivalent name for a compound/formula, equivalent when accounting for units, etc.).
Mark it as incorrect if it is not equivalent to the reference answer.

***

The problem:
\texttt{\{problem\}}

***

The reference answer:
\texttt{\{reference\_answer\}}

***

The attempted answer:
\texttt{\{answer\}}

***

First, think step-by-step about whether the attempted answer matches the reference answer.
If the attempted answer is correct, write "VERDICT: CORRECT" in the last line of your response, with no other text or formatting. If it is incorrect, write "VERDICT: INCORRECT".
\end{shadedbox}

\begin{shadedbox}{FrontierScience-Research Judge Model Prompt}
You are grading a science exam.

You will be given the problem, attempted answer, and a rubric to grade the answer. The rubric will total up to 10 points.

Evaluate the attemped answer against the provided rubric. Pay close attention to detail and grade it strictly, but fairly.
Only evaluate against the rubric, as you yourself should not make any judgements (e.g., even if you think the answer is correct but rubric is wrong, you should treat the rubric as the gold standard).
Return the absolute total number of points earned (it can be a decimal based on the rubric).
***

The problem:
\texttt{\{problem\}}

***

The rubric:
\texttt{\{rubric\}}

***

The attempted answer:
\texttt{\{answer\}}

***

First, think step-by-step about each rubric item. Explain your reasoning for each rubric item.
Then, tally the points up and write \verb|VERDICT: <total_points>| in the last line of your response, no other text. For example, \verb|VERDICT: 2.5| or \verb|VERDICT: 8|.
\end{shadedbox}

\section{Problem requirements}

We display a summarized list of requirements given to each problem writer for both the Olympiad and the Research set.

\begin{shadedbox}{Research Problem Guidelines}

\textbf{Question clarity:}
\begin{itemize}
  \item All necessary background information, variables, notation and assumptions must be explicitly defined.
  \item The model should have access to the same inputs that would be necessary for an expert in the field to solve the question. If a model fails due to missing information, adjust the description or task wording.
\end{itemize}

\textbf{Originality:}
\begin{itemize}
  \item If the question is inspired by a publicly available source (e.g., research paper), the question and/or answer values should be substantially modified.
\end{itemize}

\textbf{Grading consistency:}
\begin{itemize}
  \item Rubric items must be independent and objectively assessable.
  \item Each rubric description should:
    \begin{itemize}
      \item Be affirmative, clear, and explicitly state required conditions for credit, avoiding vagueness.
      \item Provide specific pass/fail conditions (e.g., ``Writes the following equation X'').
      \item Define all variables and acronyms.
    \end{itemize}
  \item Ensure discrepancies between human and model grading results do not exceed 0.5-points overall.
\end{itemize}

\textbf{Difficulty:}
\begin{itemize}
  \item Problems should be sufficiently challenging, typically requiring 3--5 hours to draft, to adequately test depth of reasoning.
  \item The questions should evaluate the model's ability to reason at a complex level -- researchers should test problem solving rather than prose, search or recency (knowledge cutoffs).
\end{itemize}

\end{shadedbox}

\begin{shadedbox}{Olympiad Problem Guidelines}

\textbf{Correctness -- ensure problem is correct and complete:}
\begin{itemize}
  \item Provide all assumptions and define all variables.
  \item Use correct physics/equations.
  \item Make sure the problems cannot be interpreted in multiple ways.
\end{itemize}

\textbf{Verifiable -- ensure problem is verifiable:}
\begin{itemize}
  \item Provide all variables used in the final answer.
  \item Make sure information about directions/plus/minus signs is given.
  \item Clear LaTeX formulations -- no Unicode characters and ensure all functions are formatted correctly (a common mistake is to write \texttt{sin} rather than \(\sin\)).
  \item Define a variable for what you ask for in the question (e.g., do not write ``find an expression for the energy in terms of the following variables''; instead write ``find an expression for $E$, the energy, in terms of the following variables'').
  \item Ensure the answer is a single numeric or algebraic expression.
  \item If using units in the final answer, explicitly specify the unit you want and what symbol to use (e.g., ``give your final answer in units of meters using the symbol m'').
\end{itemize}

\end{shadedbox}